\documentclass{sig-alternate}

\usepackage{amsmath}
\usepackage{algorithm}
\usepackage[noend]{algpseudocode}

\usepackage{graphicx}
\usepackage{subcaption}
\usepackage{multirow}

\makeatletter
\def\BState{\State\hskip-\ALG@thistlm}
\makeatother

\begin{document}
%
\conferenceinfo{28 Dec}{2015}

\title{Outlier Detection in Large-Scale Traffic Data By Na\"ive Bayes Method and Gaussian Mixture Model Method }

%
%
%
%
%

\numberofauthors{5} 
%
\author{
%
%
\alignauthor Philip Lam\\
       \affaddr{Department of Mathematics}\\
       \affaddr{Hong Kong Baptist University}\\
       \affaddr{Kowloon Tong, Hong Kong}\\
       \email{12015652@life.hkbu.edu.hk}
\alignauthor Lili Wang\\
       \affaddr{Department of Mathematics}\\
       \affaddr{Hong Kong Baptist University}\\
       \affaddr{Kowloon Tong, Hong Kong}\\
       \email{llwang\_hk@126.com}
\alignauthor Henry Y.T. Ngan\titlenote{This author is the corresponding author.}\\
       \affaddr{Department of Mathematics}\\
       \affaddr{Hong Kong Baptist University}\\
       \affaddr{Kowloon Tong, Hong Kong}\\
       \email{ytngan@hkbu.edu.hk}
\and  
\alignauthor Nelson H.C. Yung\\
       \affaddr{Department of Electronic \& Electrical Engineering,}\\
       \affaddr{The University of Hong Kong}\\
       \affaddr{Pokfulam, Hong Kong}\\
       \email{ypl.nyung@gmail.com}
\alignauthor Anthony G.O. Yeh\\
       \affaddr{Department of Urban Planning \& Design}\\
       \affaddr{The University of Hong Kong}\\
       \affaddr{Pokfulam, Hong Kong}\\
       \email{hdxugoy@hkucc.hku.hk}
}

\maketitle
\begin{abstract}
It is meaningful to detect outliers in traffic data for traffic management. However, this
is a massive task for people from large-scale database to distinguish outliers. In this paper,
we present two methods: Kernel Smoothing Na\"ive Bayes (NB) method and Gaussian Mixture Model
(GMM) method to automatically detect any hardware errors as well as abnormal traffic events in
traffic data collected at a four-arm junction in Hong Kong. Traffic data was recorded in a video
format, and converted to spatial-temporal (ST) traffic signals by statistics. The ST signals are
then projected to a two-dimensional (2D) (x, y)-coordinate plane by Principal Component Analysis
(PCA) for dimension reduction. We assume that inlier data are normal distributed. As such, the NB
and GMM methods are successfully applied in OD (Outlier Detection) for traffic data. The kernel
smooth NB method assumes the existence of kernel distributions in traffic data and uses Bayes'
Theorem to perform OD. In contrast, the GMM method believes the traffic data is formed by the
mixture of Gaussian distributions and exploits confidence region for OD. This paper would address
the modeling of each method and evaluate their respective performances. \ Experimental results
show that the NB algorithm with Triangle kernel and GMM method achieve up to \textbf{93.78\%} and \textbf{94.50\%}
accuracies, respectively.
\end{abstract}


\terms{Outlier detection}

\keywords{large-scale,
traffic data,
Naive Bayes,
Gaussian Mixture Model}

\section{Introduction}
It is desired that most events to be safe, stable or rather predictable. Identifying an unusual
event is a typical and vital topic in different fields, such as intrusion detection in cyber security,
fraud detection for credit cards, insurance or health care and fault detection in safety critical systems
\cite{Kumar}. We may call this unusual event as outlier or abnormality which is different from a usual event.
Outlier is usually minor in a group of events/data while inlier is the majority.

OD refers to detect any abnormal element in data which is not consistent with an expected behavior \cite{Hodge}.
A good OD method should be accurate to detect outlier and less erroneous judgment on inlier data. Many
OD methods \cite{Kriegel} have been developed in recent years. In \cite{Hodge},\cite{Kumar},  nearest neighbor approach was proposed
to consider the distance or the similarity between two data instances. The assumption behind the approach
is that inliers in data should be dense and outlier(s) is far from these dense inliers.

NB classifiers were developed in the 1950s. The NB classifiers are constructed based on Bayes' theorem, and are widely
used for text categorization with superior performance \cite{Russell}. Another popular application is spam filtering
among numerical emails \cite{Sahami}. The kernel smooth NB method in this paper supposes the traffic data could be
modeled by kernel distributions so that the Bayes' theorem is applied. In contrast, the GMM method presumed
a mixture of Gaussian distributions and confidence region is employed . GMM, firstly introduced by Aitkin
and Wilson (1980) \cite{Aitkin}, is \ an appropriate way of handle data with multiple outliers \cite{Scott}. The GMM method
is useful in dealing with speech recognition \cite{Stuttle}.

\begin{figure}[ht]
  \begin{subfigure}[b]{0.5\linewidth}
    \centering
    \includegraphics[width=1\linewidth]{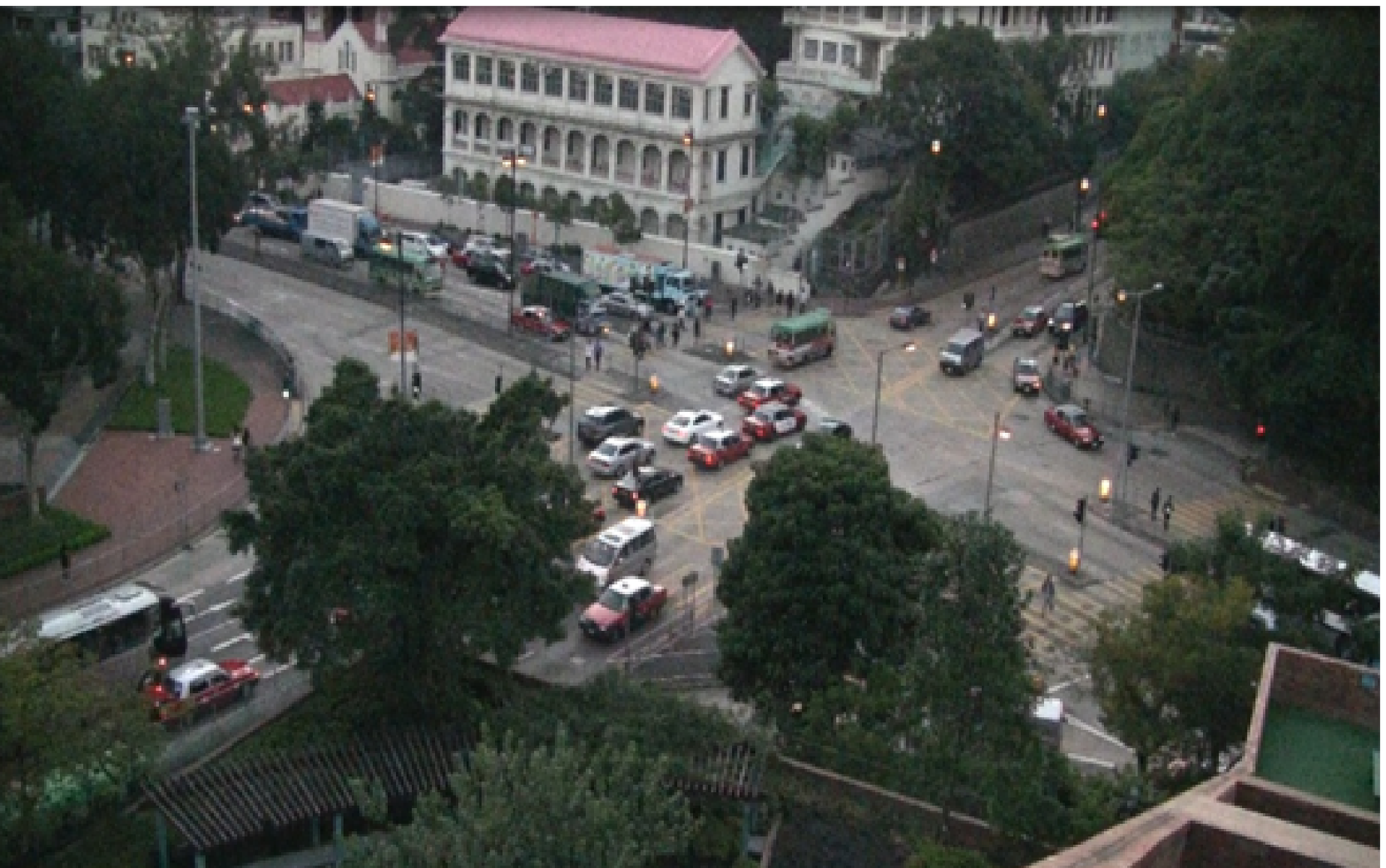}
    \caption{The real scene}
    \label{fig1:a}
    \vspace{4ex}
  \end{subfigure}
  \begin{subfigure}[b]{0.5\linewidth}
    \centering
    \includegraphics[width=1\linewidth]{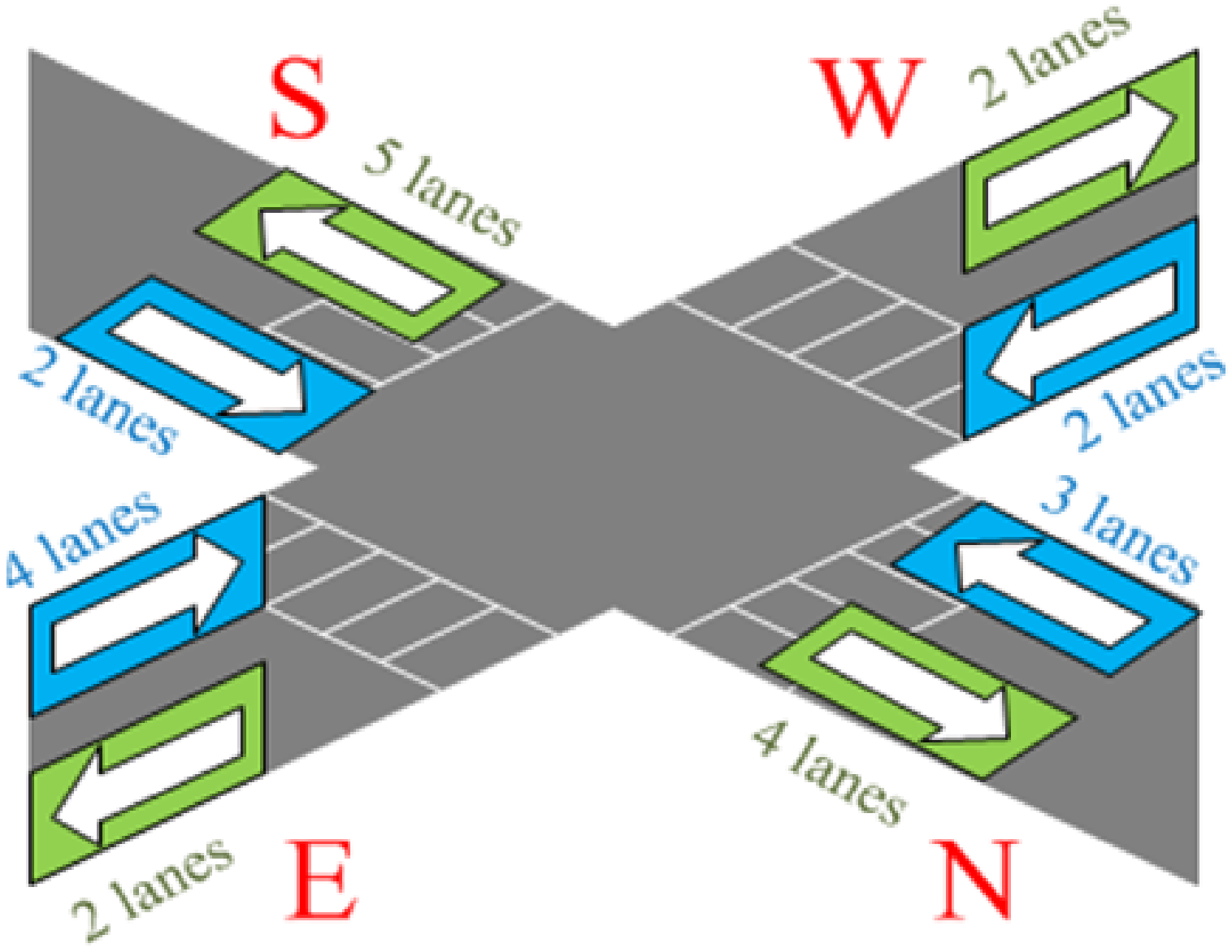}
    \caption{4-arm junction}
    \label{fig1:b}
    \vspace{4ex}
  \end{subfigure}
  \begin{subfigure}[b]{0.5\linewidth}
    \centering
    \includegraphics[width=1\linewidth]{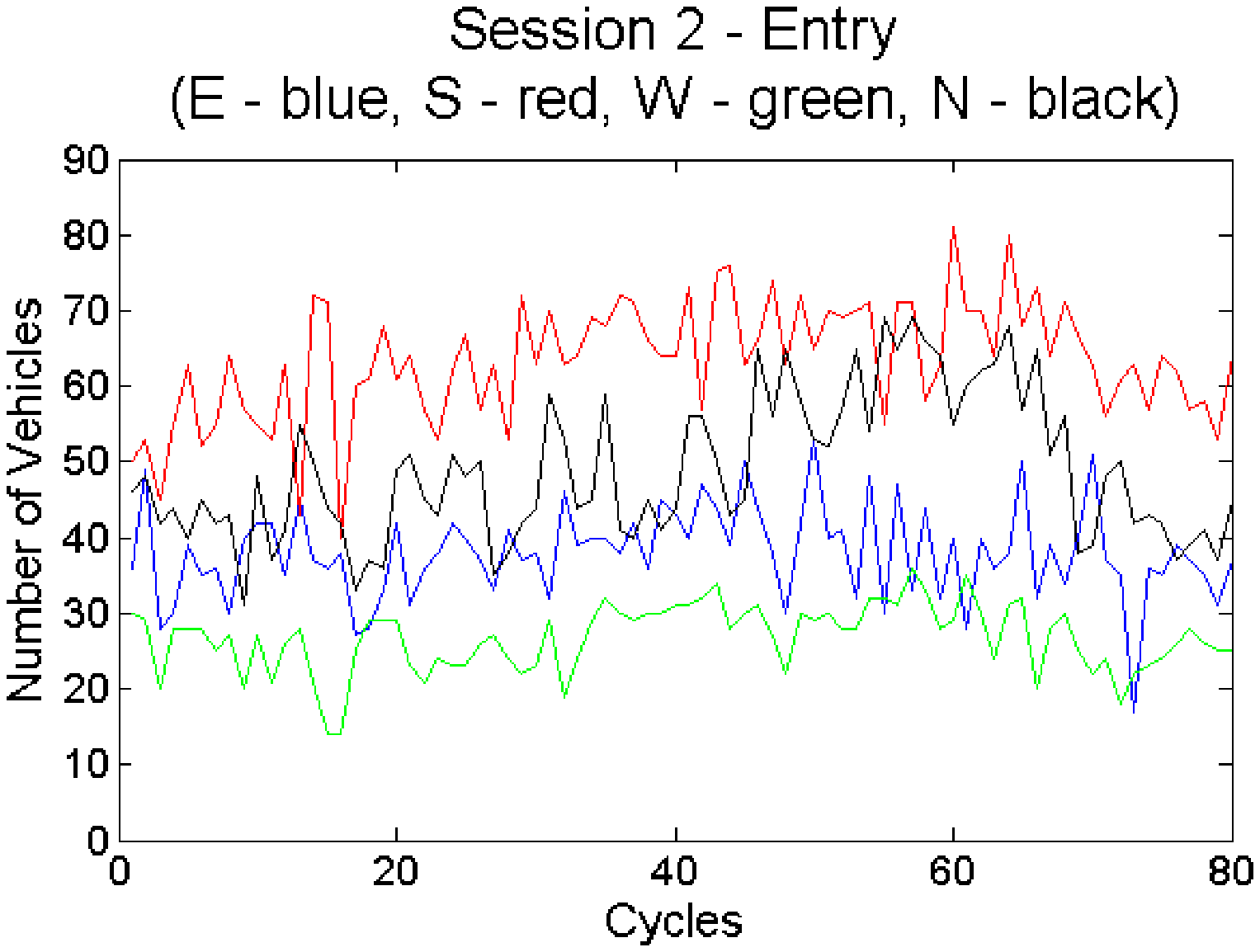}
    \caption{Session 2 Entry: All signals are normal }
    \label{fig1:c}
  \end{subfigure}
  \begin{subfigure}[b]{0.5\linewidth}
    \centering
    \includegraphics[width=1\linewidth]{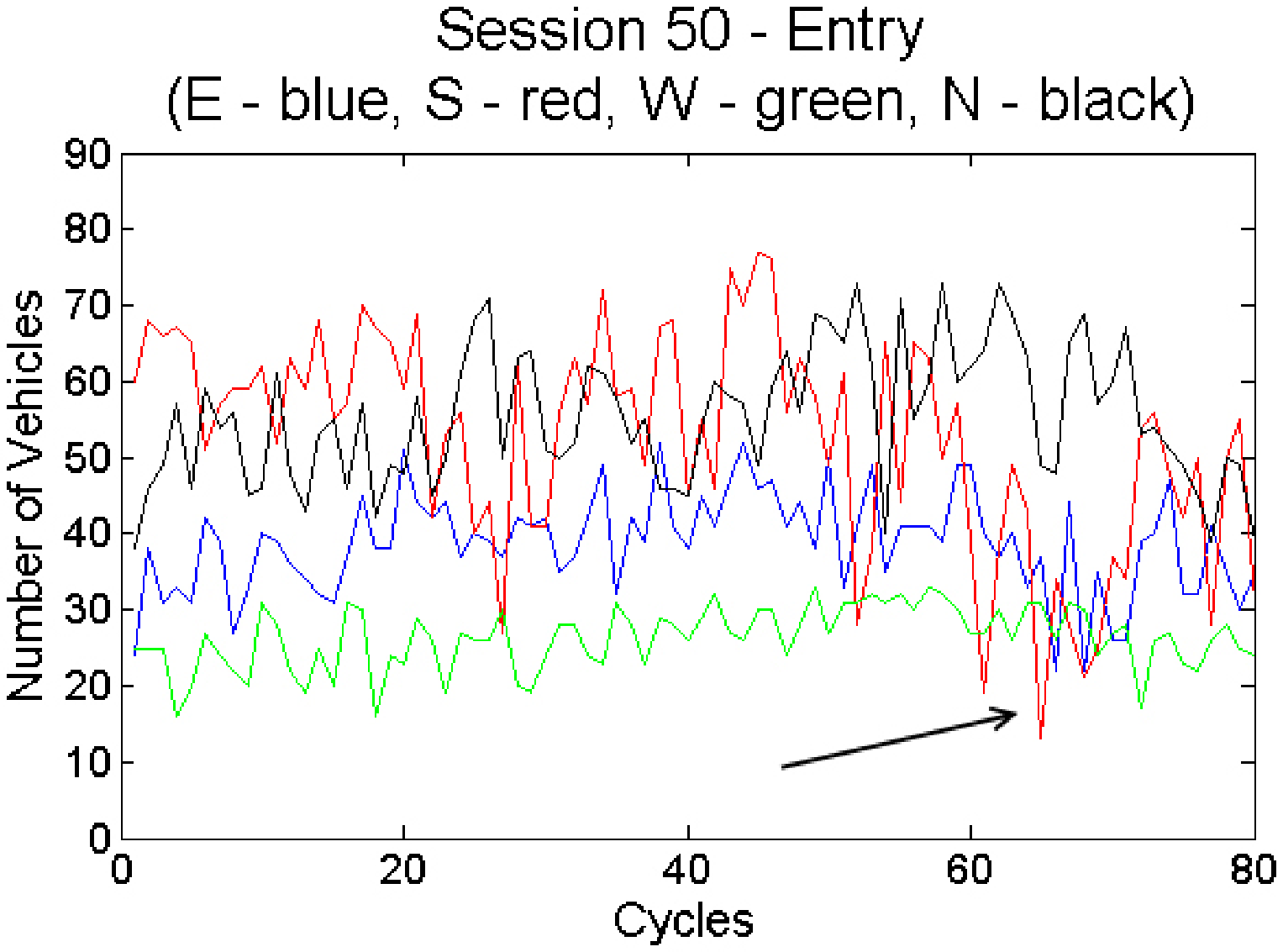}
    \caption{Session 50: Entries E,W,N are normal, Entry S is abnormal }
    \label{fig1:d}
  \end{subfigure}
  \caption{(a) A generic diagram of the 4-arm junction; (b) sample of the real scene; (c) normal ST signals; (d) abnormal ST signals (with arrow). }
  \label{fig1}
\end{figure}

For traffic data, detecting anomaly traffic event would be better to deal with the traffic problems. This
research aims at detecting outliers for a large-scale traffic database from Hong Kong. The original video
was taken at a four-arm junction as shown in Fig. \ref{fig1:a}. The 4-arm junction can be expressed as an ideal map
like Fig. \ref{fig1:b}.
In total, the traffic data for 31 days was recorded
with two sessions per day: AM (07:00-10:00) and PM (17:00-20:00). For each session, the original video data
are dividing into 19 spatial temporal (ST) signals. Fig. \ref{fig1:c} demonstrates an sample of 4 normal ST signals.
These ST signals have different number of traffic cycles
because smoother traffic flow would lead to a shorter traffic cycle and result in more traffic cycles in
one session. Also, the ST signals in each direction suffer a high degree of similarities (Fig. \ref{fig1:d})
among each other
or across signals in different traffic direction. Therefore, it requires a truncation process (to standardize
the length of a signal's cycles) and a signal domain transformation (to remedy the signal similarities).
Herein, each ST signal was processed by principal component analysis (PCA) \cite{Ngan12} to reduce the dimension of
signal representation. Then, an OD method was carried out based on the PCA-processed (x,y)-coordinates plane.

In this paper, a detail investigation of NB and GMM classifiers in large-scale traffic data is carried out,
for these two methods are newly applied in traffic OD. Experimental results demonstrate that the NB and GMM
methods can achieve 93.78\% and 94.50\% detection success rates (DSRs), respectively. These performance is
comparable to previous evaluation on other OD methods \cite{Ngan15}, \cite{Ngan15b} including Gaussian mixture model (80.86\%),
one-class SVM (59.27\%), S-estimator (76.20\%) and kernel density estimation (95.20\%). \

The organization of this paper is as follows: Section II gives a review of related work about OD. Sections III
and IV present details of two proposed OD methods and their experimental results, and Section V concludes this paper.

\section{Related Work}

The popular OD methods include statistical, nearest neighborhood, spectral approaches and learning based
approaches. Details about these methods are given as follows.

\subsection{Statistical Approach}
Statistical approach is one of the earliest approaches in OD.
It assumes that all inliers occur in a high probability region of this distribution model
when outliers deviate strongly from the inliers. Statistical OD methods can perform parametric
techniques like GMM or regression, or non-parametric techniques like histogram based or kernel
function based. This approach would perform well if data distribution is assumed well. However,
if the data distribution assumption is false, the result would be far away from the correct situation \cite{Chandola}, \cite{Kriegel}.

\subsection{Nearest Neighbor Approach}
Nearest neighbor approach considers the distance or the similarity between two data instances. The assumption
behind the approach is that inliers in data should be dense and outlier(s) is far from these dense inliers
\cite{Kumar}, \cite{Hodge}. This approach can be easily realized in an unsupervised way that could require high computational
complexity. As a result, it is hard to deal with very complex data \cite{Kriegel}.

There are two major variant methods related to this approach, density measurement based methods, such as Local
Outlier Factor (LOF) \cite{Yoon}, Influenced Outlierness (INFLO) \cite{Chandola} and Local Outlier Correlation Integral (LOCI)
\cite{Chandola}, or distance measurement based methods, like Mahalanobis Distance \cite{Jayakumar}, DB($\varepsilon,\pi$)-Outliers,
Outlier Scoring based on k-NN distances \cite{Hautamaki}, Resolution-based Outlier Factor (ROF) \cite{Kriegel}.

\subsection{Spectral Approach}
Spectral approach suggests that inliers and outliers could appear significantly different in a spectral domain.
The approach detects outliers by embedding data into a lower dimensional subspace. The approach is useless if
inliers and outliers in data are not separable in the lower dimensional subspace. In addition, it requires high
computational complexity \cite{Kriegel}.

\subsection{Learning Approach}
Learning approach is using various training methodologies \cite{Minsky} (or so-called trained machines) to train the
input data. The related OD methods are neural networks, na\"ive Bayesian network, support vector machines
(SVM). The assumption is that inliers and outliers in data can be easily distinguished by the trained machines.
Therefore, the machines can test and classify a test instance into either an inlier group or an outlier group
in OD \cite{Kriegel}. There are many algorithms that can be used in multi-learning class approach, and they have a faster
testing phase than other approach. However, accurate labeling for various normal classes is often impossible \cite{Kriegel}.

In this paper, an investigation of NB classifier and GMM classifier to model the PCA-processed (x,y)-data and
their performance for OD in traffic data would be carried out.

\section{Kernel Smoothing NB Method}
NB classifier was developed in 1950s for text retrieval \cite{Stuttle}. It is also widely used for spam filtering
among numerical emails \cite{Aggarwal}.

\subsection{NB Classifier}
NB classifier \cite{Chandola} is developed based on Bayes' Theorem which takes the form of
\begin{eqnarray}
P(H\mid E)=\frac{P(H)P(E\mid H)}{P(E)}
\end{eqnarray}
Based on the probability P(E),P(H), and the conditional probability P(E{\textbar}H), the posteriori probability P(H{\textbar}E),
which denotes the possibility of event H conditioned on an occurred event E, can be obtained. Naive Bayes
classifiers are based on the information of the training data, and then determined the highest possible class
of testing data from their information.

\subsection{Kernel Distribution}
Kernel Distribution is a non-parametric distribution to estimate each training point as some independent
distributions into the whole distribution \cite{John}.
\begin{eqnarray}
\hat{f_h}=\frac{1}{n}{\sum\limits_{i=1}^n K_h(x-x_i)}
\end{eqnarray}
Kernel Distribution is estimated by \ sample data points with a kernel density. The commonly used
kernels are listed as follows:\\
Box kernel:
\begin{eqnarray}
K(x)=0.5I\{ \left|{x}\right|\leq 1\}
\end{eqnarray}
Triangle kernel:
\begin{eqnarray}
K(x)=(1-\left|{x}\right|)I\{ \left|{x}\right|\leq 1\}
\end{eqnarray}
Epanechnikov kernel:
\begin{eqnarray}
K(x)=0.75(1-x^2)I\{ \left|{x}\right|\leq 1\}
\end{eqnarray}
Normal kernel:
\begin{eqnarray}
K(x)=\frac{1}{2\pi} exp(-0.5x^2)
\end{eqnarray}
where \textit{x} is replaced by \textit{y} for y-axis kernel.

\subsection{Kernel Smoothing NB classifier for OD}
Kernel smoothing NB (KSNB) classifier is a trained Kernel Distribution based classifier. Different
kernels with different widths for each predictor or class are available in KSNB classifier. We use the
trained KSNB classifiers to detect outliers, where the classifiers would automatically set a bandwidth
value for each feature and class which is optimal for a Gaussian model. Then, a corresponding region
from contours surrounding various kernel distributions is formed an inlier region E, therefore any data
out of that region is classified as an outlier. In this method, box, triangle, Epanechnikov and normal
kernels are used for OD. The OD procedures based on KSNB classifier are given as follows:\\

\begin{figure}
\centering
\includegraphics[width=3.4in, height=3.8in]{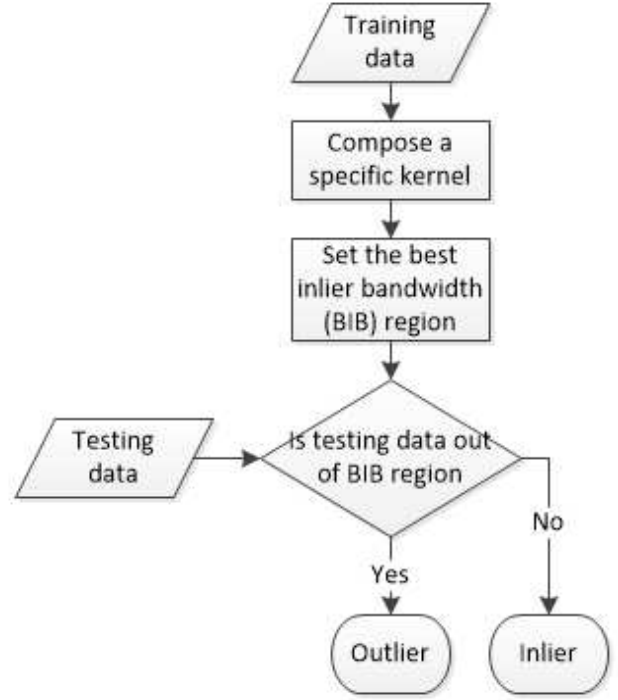}
\caption{Flowchart of the proposed NB method.}\label{FigNBflowchart}
\end{figure}

Fig. \ref{FigNBflowchart} illustrates the flowchart of the proposed NB based method, which includes a training stage
and a testing stage. The details are as below.

\paragraph{A. Training Stage}
\noindent Step 1: Feed training data (all inliers) into NB classifiers.\\
Step 2: Fit training data into the specified kernel distributions defined in Eqns. (3-6).\\
Step 3: Combine each kernel distribution into a whole distribution.\\
Step 4: Set a bandwidth (BIB) of the inlier region automatically from the kernel distribution.

\paragraph{B. Testing Stage}
\noindent Step 1: Input testing data into the trained NB classifiers.\\
Step 2: Data label as outlier if the data is out of the inlier region.\\
Step 3: Output result.\\

\noindent In addition, \textbf{Algorithm \ref{NBalg}} is provided below.

\begin{algorithm}
\caption{NB classifier with different kernel distribution}\label{NBalg}
\begin{algorithmic}[1]
\Require~~ The inliers data set $\mathbf{A}=(x_i,y_i)$ and testing data set $\mathbf{B}=(x_i,y_i)$
\Procedure {Employ Kernel Distribution}{}
\State {define outliers set} $\mathbf{H}$
\State $\mathbf{for}$ {every data point} $\mathbf{A}$
\State {form in kernel distribution} $(K_i(x-x_i),K_i(y-y_i))$
\State $\mathbf{end}$
\State $\hat{f_h}=\frac{1}{n}{\sum\limits_{i=1}^n K_h(x-x_i)}$
\State {set the bandwidth value by} $\hat{f_h}$
\State {form an inlier region} $E$
\State {for all} $i$ {in} $\mathbf{B}$
\If {$(x_i,y_i)\in E$}
\State $(x_i,y_i)\in \sim H$
\Else $(x_i,y_i)\in H$
\State $\mathbf{end}$
\EndIf
\State {output labels}
\EndProcedure
\end{algorithmic}
\end{algorithm}

\subsection{Experimental results}
Experimental results based on the kernel smoothing NB method are listed in Table \ref{diffNBkernelRes}. Accuracy (Acc.),
positive predictive value (PPV), negative predictive value (NPV), sensitivity (Sen.), false positive
rate (FPR) are employed as measurement metrics and their definitions can be referred to \cite{Ngan15}. From the
results, we can see that overall accuracy among all kernels is almost higher than \textbf{90\%}. Overall
NPV among all kernels is higher than \textbf{95\%.}

\begin{table}
   \caption{Performance of the NB Method Among Different Kernels.}
\begin{center}
\begin{tabular}{|c|c|c|c|c|c|}
   \hline \hline
    & Kernel & Gaussian & Box & Epanechnikov & Triangle\\ \hline \hline
    \multirow{5}{*}{AM} &{Acc.} &97.13 &91.39 &95.22 &96.17\\
    &{PPV}	&{NA}	&18.18	&40.00	&50.00\\
    &{NPV}	&97.13	&97.93	&98.09	&98.09\\
    &{Sen.}	&0.00	&33.33 	&33.33	&33.33\\
    &{FPR}	&0.00	&6.84	&2.87	&1.91\\ \hline
    \multirow{5}{*}{PM} &{Acc.} &91.87	&85.65	&89.47	&91.39\\
    &{PPV}	&100.00	&22.71	&29.49	&36.81\\
    &{NPV}	&91.65	&96.21	&96.49	&96.49\\
    &{Sen.}	&5.00	&55.50	&55.50	&55.50\\
    &{FPR}	&0.00	&12.60	&8.31	&6.30\\ \hline
    \multirow{5}{*}{All} &{Acc.} &94.50	&88.52	&92.345	&93.78\\
    &{PPV}	&100.00	&20.45	&34.75	&43.41\\
    &{NPV}	&94.39	&97.07	&97.29	&97.29\\
    &{Sen.}	&2.50	&44.42	&44.42	&44.42\\
    &{FPR}	&0.00	&9.72	&5.59	&4.11\\ \hline
   \hline
\end{tabular}
\end{center} \label{diffNBkernelRes}
\end{table}

\begin{figure}
\centering
\includegraphics[width=3.3in, height=2.2in]{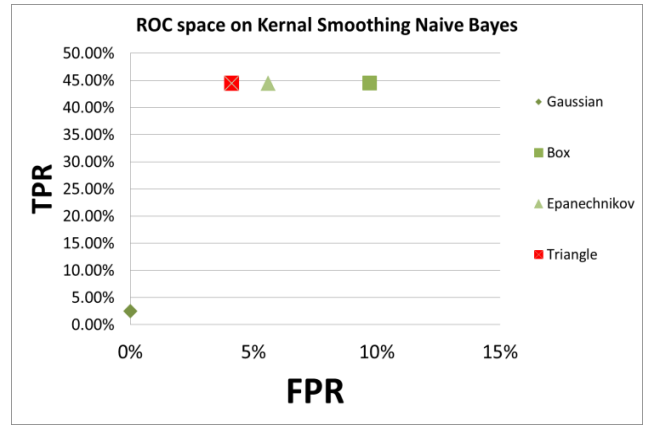}
\caption{ROC plot for the NB method based on four kernels.}\label{FigNB-ROC}
\end{figure}

The OD scatters corresponding to the four kernels are shown in Fig. \ref{FigNB-ROC}. From the figure, one observes that
the scatter based on Triangle kernel is closest to the perfect classification point. Among the NB method
with different kernels, Gaussian kernel usually has better performance. However, in our evaluation, the OD
performance based on NB method with Gaussian kernel is very close to the random-guess line ( black line)
as shown in ROC space of Fig. \ref{FigNB-ROC}. It is just a little better than random guessing that all testing data are
inliers. Therefore, Gaussian kernel is found to be not a good kernel in this method for OD.\textbf{ }For
other three kernels listed in Table I, the triangle kernel shows the best performance that is closest to
the perfect classification point with the highest accuracy \textbf{(93.78\%)}. For AM or PM OD, the NB method
with the triangle kernel still outperforms other three kernels That may indicate that the NB method with a
triangle kernel is significantly good to describe the behaviors of PCA traffic data.\\

\section{GMM Method}

The GMM method was firstly introduced in \cite{Aitkin}, the mixture model as a way of handle
data with multiple outliers \cite{Hautamaki}. The GMM method is good in deal with speech recognition \cite{Stuttle}.

\subsection{Gaussian Distribution}
In mathematics, a one-dimensional \textbf{Gaussian distribution} is a distribution function of the possibility
density form:\\
\begin{eqnarray}
f(x)=\frac{1}{\sigma \sqrt{2\pi}}e^{- \frac{1}{2} (\frac{x-M}{\sigma}) ^2}
\end{eqnarray}
where $x\sim N(M,\sigma^2)$, $M$ denotes the expectation of $x$, $\sigma^2$ denotes the variance of $x$.
In term of a two-dimensional expression, the Gaussian distribution is
\begin{eqnarray}
f(x,y)=\frac{1}{2\pi \sigma^2 }e^{- \frac{1}{2} ( \frac{x-M_x}{\sigma_x} + \frac{y-M_y}{\sigma_y}) ^2}
\end{eqnarray}
where $x\sim N(M_x,\sigma_x^2)$ and $y\sim N(M_y,\sigma_y^2)$ .

\subsection{GMM}
\textbf{GMM} is a parametric probability density function represented as a weighted sum of Gaussian component
densities. The p.d.f. of the mixture model among $f_i$ is $F(x,y)$ which takes the form of
\begin{eqnarray}
F(x_i,y_i)= \sum\limits_{i=1}^n w_i f_i (x_i,y_i)
\end{eqnarray}
where $w_i$ is the mixture weight \cite{Reynolds}.

\subsection{Methodology}
For OD in this section, we will use confidence region (CR) with a Bonferroni adjustment
as criterion in \ significant level. Herein, $\alpha =0.1, 0.05, 0.01$.

For a CR with Bonferroni adjustment, GMM CR for X is
\begin{eqnarray}
[\bar{x}-t_\frac{\alpha}{2*3} \sqrt{var(x)},\bar{x}+t_\frac{\alpha}{2*3} \sqrt{var(x)} ]
\end{eqnarray}
GMM confidence region for \textbf{Y }is
\begin{eqnarray}
[\bar{y}-t_\frac{\alpha}{2*3} \sqrt{var(y)},\bar{y}+t_\frac{\alpha}{2*3} \sqrt{var(y)} ]
\end{eqnarray}
where $t_k$ is student's value in $k$ significant level.
The GMM method models fit the training data by using several Gaussian distributions.

In this section, data fitting will use the expectation maximization (EM) algorithm \cite{Reynolds},
which assigns posterior probabilities to each component density with respect to each observation,
to get the best $w_i$ for each $i$ in (9).

\subsection{Procedure}

\begin{figure}
\centering
\includegraphics[width=3.4in, height=3.9in]{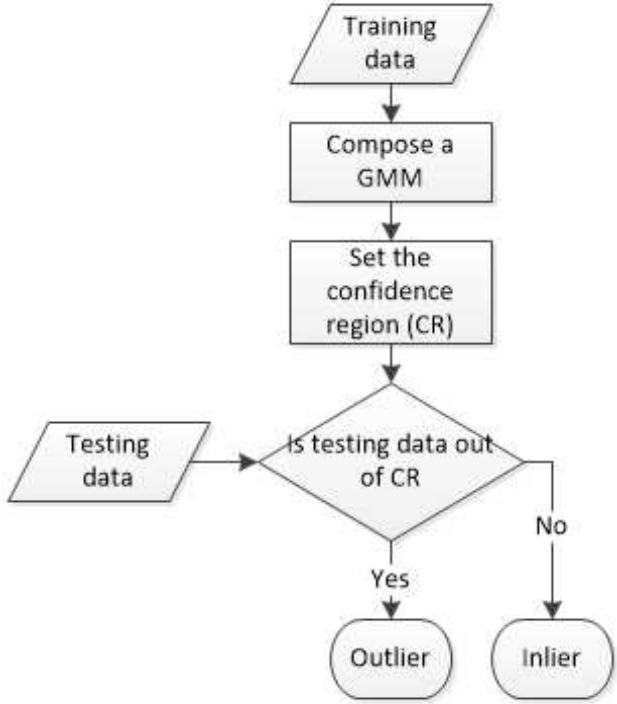}
\caption{Flowchart of the proposed GMM method.}\label{FigGMMflowchart}
\end{figure}

A flowchart of the proposed GMM method, including a training stage and a testing stage is
presented in Fig. \ref{FigGMMflowchart}. The details are as below.

\paragraph{A. Training Stage}
\noindent Step 1: Input inlier training data.\\
Step 2: Fit each data into individual identical Gaussian distribution model.\\
Step 3: Mixture all models into a whole GMM model (9).\\
Step 4: Set the criteria region for different criterion method and different significance
level [(i) 0.10, (ii) 0.05, (iii) 0.01].

\paragraph{B. Testing Stage}
\noindent Step 1: Input testing data set.\\
Step 2: Label data as an outlier if the data is out of the criteria region \\

\noindent Also, \textbf{Algorithm \ref{GMMalg}} is given.

\begin{algorithm}
\caption{GMM with 1 mixture classifier in confidence region method}\label{GMMalg}
\begin{algorithmic}[1]
\Require~~ The inliers data set $\mathbf{A}=(x_i,y_i)$ and testing data set $\mathbf{B}=(x_i,y_i)$
\Procedure {Employ Confidence Region}{}
\State {define outliers set} $\mathbf{H}$
\State $\mathbf{for}$ {every data point} $\mathbf{A}$
\State {form a Gaussian model with} \\
       $\mathbf{GMM} \sim N \left( (\bar{x},\bar{y})^T, \begin{pmatrix}
          var(x)& cov(x,y)\\
          cov(x,y) & var (y)\\
          \end{pmatrix} \right)$
\State $\mathbf{end}$
\State {fixed the significant level} $\alpha$
\State {construct \textbf{GMM} \textbf{x} confidence region \textbf{X} with:}
\State $[\bar{x}-t_\frac{\alpha}{2*3} \sqrt{var(x)},\bar{x}+t_\frac{\alpha}{2*3} \sqrt{var(x)} ] $
\State {construct \textbf{GMM} \textbf{y }confidence region \textbf{Y }with:}
\State $[\bar{y}-t_\frac{\alpha}{2*3} \sqrt{var(y)},\bar{y}+t_\frac{\alpha}{2*3} \sqrt{var(y)} ] $
\State {for all} $i$ {in} $\mathbf{B}$
\If {$(x_i)\in X$}
\If {$(y_i)\in Y$}
\State $(x_i,y_i)\in \sim H$
\Else $(x_i,y_i)\in H$
\State $\mathbf{end}$
\EndIf
\Else $(x_i,y_i)\in H$
\State $\mathbf{end}$
\EndIf
\State {output results}
\EndProcedure
\end{algorithmic}
\end{algorithm}

\subsection{Experimental results}
\begin{figure}
\centering
\includegraphics[width=3.3in, height=2.2in]{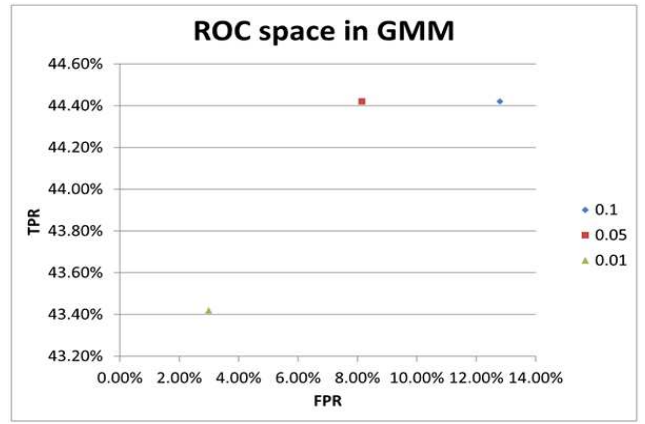}
\caption{ROC plot for the GMM method based on four kernels.}\label{FigGMM-ROC}
\end{figure}

Experimental results based on the GMM method are listed in Table \ref{diffGMM}. The performances in terms of FPR,
PPV and Acc. are gradually better with the decrease of significant level. The highest accuracies in both
AM (96.65\%) and PM (94.50\%) sessions are achieved when the significant level is 0.01. A ROC plot on the
GMM method is shown in Fig. \ref{FigGMM-ROC}. By observation, CR in the 0.01 level is the closest point to
the perfect
classification point. Inlier data points lying in the centroid of the GMM model and the CR cover at least
$\alpha$\ significant level built by training data set. Therefore, the CR criterion is good at describing
the inlier group which is centered.

\begin{table}
   \caption{Performance of the GMM Method in Various Significant Levels.}
\begin{center}
\begin{tabular}{|c|c|c|c|c|}
   \hline \hline
    Level &  & 0.10 & 0.05 & 0.01\\ \hline \hline
    \multirow{5}{*}{AM} &{Acc.} &89.47 &93.78 &96.65\\
    &{PPV}	&9.09	&14.29 	&50.00\\
    &{NPV}	&97.70 &97.99 &98.09\\
    &{Sen.}	&33.33 &33.33 &33.33\\
    &{FPR}	&9.04 &4.50 &1.44 \\ \hline
    \multirow{5}{*}{PM} &{Acc.} &82.30 &86.12 &92.34\\
    &{PPV}	&18.83 	&28.02 &46.30\\
    &{NPV}	&96.26 &96.52 &96.11\\
    &{Sen.}	&55.50 &55.50 &53.50\\
    &{FPR}	&16.55 &11.81 &4.55\\ \hline
    \multirow{5}{*}{All} &{Acc.} &85.89 &89.95 &94.50\\
    &{PPV}	&13.96 &21.16 &48.15\\
    &{NPV}	&96.98 &97.26 &97.10\\
    &{Sen.}	&44.42 &44.42 &43.42\\
    &{FPR}	&12.80 &8.16 &3.00\\ \hline
   \hline
\end{tabular}
\end{center} \label{diffGMM}
\end{table}

\section{Conclusion}
In this paper, we present two OD methods, kernel smoothing NB and GMM, to detect outliers in large-scale
traffic data. The kernel smoothing NB method utilizes the trained kernel smoothing NB classifiers to detect
outliers, in which the classifiers set the best bandwidth value with the inlier region. Any data points out
of that region are classified as outliers. In the GMM method, the rectangular confidence region is formed
in Bonferroni adjustment way. True $\alpha$\ significant level region of the GMM can be constructed as
the inlier region for the more accuracy. Experimental results show that the two algorithms can achieve pleasing
detection accuracies compared with the OD methods in our previous studies \cite{Ngan15}, \cite{Ngan15b}, including Gaussian mixture
model, one-class SVM, S-estimator and kernel density estimation.

\section{Acknowledgments}
This research is supported by Hong Kong RGC GRF: 12201814 and HKBU FRG2/14-15/075. \



\begin{thebibliography}{1}

\bibitem{Aggarwal} C. Aggarwal, {\it Outlier Analysis}, Springer, 2013.

\bibitem{Aitkin} M. Aitkin and G.T. Wilson, ``Mixture models, outliers, and the EM algorithm," {\it Technometrics}, 22, 325-331, 1980.


\bibitem{Chandola} V. Chandola, A. Banerjee, and V. Kumar, ``Anomaly detection: A survey," {\it ACM Computing Surveys}, 41(3), pp.1-58, 2009.

\bibitem{Hautamaki} V. Hautamaki , I. Karkkainen and P. Franti, ``Outlier Detection Using k-Nearest Neighbour Graph," {\it Proc. IEEE ICPR},  vol. 3, pp.430-433, 2004.

\bibitem{Hodge} V. J. Hodge, J. Austin, ``A Survey of Outlier Detection Methodologies," {\it Artificial Intelligence Review}, vol. 22 no. 2, pp.85-126, 2004.

\bibitem{Jayakumar} G.S.D.S. Jayakumar and B. J. Thomas ``A New Procedure of Clustering Based on Multivariate Outlier Detection", {\it Science}, 11, pp. 69-84, 2013.

\bibitem{John} G.H. John and P. Langley, ``Estimating continuous distributions in Bayesian classifiers," {\it In: Hanks, P. B. eds., Proc. 11th Conf. Uncertainty in Artificial Intelligence}, Morgan Kaufmann, San Francisco, CA, pp. 338-345, 1995.

\bibitem{Kriegel} H. Kriegel, P. Kroger and A. Zimek, ``Outlier Detection Techniques," {\it Tutorial at the 13th Pacific-Asia Conference on Knowledge Discovery and Data Mining}, 2009.

\bibitem{Kumar} V. Kumar, ``Parallel and Distributed Computing of Cybersecurity," {\it IEEE Distributed Systems Online}, vol. 6, no. 10, 2005.

\bibitem{Minsky} M. Minsky and S. Papert, ``An Introduction to Computational Geometry," {\it MIT Press}, ISBN 0-262-63022-2, 1969.

\bibitem{Ngan12} H.Y.T. Ngan, N.H.C. Yung and A.G.O. Yeh, ``Detection of Outliers in Traffic Data based on Dirichlet Process Mixture Model," {\it Proc. IEEE CASE}, pp 224-229, 2012.

\bibitem{Ngan15} H.Y.T. Ngan, N.H.C. Yung and A.G.O. Yeh, ``Detection of Outliers in Traffic Data based on Dirichlet Process Mixture Model," {\it IET Intelligent Transportation Systems}, vol. 9, no. 7, pp. 773-781, 2015.

\bibitem{Ngan15b} H.Y.T. Ngan, N.H.C. Yung and A.G.O. Yeh, ``A Comparative Study of Outlier Detection for Large-scale Traffic Data by One-class SVM and Kernel Density Estimation," {\it IS\&T/SPIE Electronic Imaging}, 94050I-94050I-10, 2015.

\bibitem{Reynolds} D.A. Reynolds, ``Gaussian Mixture Models," {\it Encyclopedia of Biometric Recognition}, Springer, Heidelberg, 2008.

\bibitem{Russell} S. Russell and P. Norvig, {\it Artificial Intelligence: A Modern Approach}, 2nd ed., Prentice Hall, ISBN 978--0137903955, 2003.

\bibitem{Sahami} M. Sahami, S. Dumais, D. Heckerman and E. Horvitz, ``A Bayesian approach to filtering junk e-mail," {\it AAAI Workshop on Learning for Text Categorization}, AAAI Technical Report, WS-98-05, 1998.

\bibitem{Scott} D.W. Scott, ``Outlier detection and clustering by partial mixture modelling," {\it COMPSTAT 2004 Symposium}, pp.  453-465, Heidelberg, 2005.

\bibitem{Stuttle} M. N. Stuttle, ``A Gaussian mixture model spectral representation for speech recognition," {\it Ph.D. dissertation, University of Cambridge}, 2003.

\bibitem{Yoon} K-A. Yoon, O-S. Kwon, and D-H. Bae, ``An Approach to Outlier Detection of Software Measurement Data Using the K-means Clustering Method," {\it IEEE 1st Int'l Sym. ESEM}, pp. 443-445 , 2007.

\end{thebibliography}
\end{document}